\documentclass{article}
\usepackage{spconf,amsmath,epsfig}
\usepackage{comment}
\usepackage{amsthm}%
\usepackage{multicol}%
\usepackage{mathrsfs}%
\usepackage{multirow}
\usepackage[title]{appendix}%
\usepackage{xcolor}%
\usepackage{fancyhdr}
\usepackage{textcomp}%
\usepackage{manyfoot}%
\usepackage{booktabs}%
\usepackage{algorithm}%
\usepackage{algorithmicx}%
\usepackage{algpseudocode}%
\usepackage{listings}%
\usepackage{tabularx}
\usepackage{graphicx}
\usepackage{amssymb,amsfonts}%
\usepackage{adjustbox}
\newtheorem{definition}{Definition}%

\title{HARNESSING FEATURE CLUSTERING FOR ENHANCED ANOMALY DETECTION WITH VARIATIONAL AUTOENCODER AND DYNAMIC THRESHOLD }
\twoauthors
  {Tolulope Ale, Vandana P. Janeja \sthanks{Corresponding Author.}}
   {University of Maryland Baltimore County\\
        Information Systems\\
        1000 Hilltop Circle, Baltimore, MD 21250}
  {Nicole-Jeanne Schlegel}
   {   National Oceanic and Atmospheric\\ Administration\\ Geophysical Fluid Dynamics Laboratory\\ Princeton, NJ USA}

\begin{document}
\maketitle
\begingroup
\renewcommand{\thefootnote}{}
\footnotetext{2024 IEEE International Geoscience and Remote Sensing Symposium, IGARSS 2024, 07-12 July 2024, Athens, Greece.}
\endgroup

 \vspace{-0.2in}   
\begin{abstract}
We introduce an anomaly detection method for multivariate time series data with the aim of identifying critical periods and features influencing extreme climate events like snowmelt in the Arctic. This method leverages Variational Autoencoder (VAE) integrated with dynamic thresholding and correlation-based feature clustering. This framework enhances the VAE's ability to identify localized dependencies and learn the temporal relationships in climate data, thereby improving the detection of anomalies as demonstrated by its higher F1-score on benchmark datasets. The study's main contributions include the development of a robust anomaly detection method, improving feature representation within VAEs through clustering, and creating a dynamic threshold algorithm for localized anomaly detection. This method offers explainability of climate anomalies across different regions.
\end{abstract}

\vspace{-0.2in}
\section{Introduction}
\label{sec:intro}
\vspace{-0.1in}
The Arctic is warming at a rate twice the rest of the globe; this has led to changes in the seasonal melt patterns with an increase in frequency, intensity, and spatial coverage of the melt. This extreme melting is a result of complex interactions of multiple climate features. Multivariate time series anomaly detection provides a way to study how melt patterns are changing. Understanding why a particular period is an anomaly can be more difficult when many variables are involved. As a result, introducing a robust preprocessing and postprocessing methodology might enhance the anomaly detection accuracy.

Our research is motivated by the need to develop a robust framework that captures the intricate relationship among climate features in the Arctic. This is crucial to understanding the features critical to extreme events such as snowmelt in the Arctic and its subsequent impact on climate change. For instance, the 2019 melt was exacerbated by a series of anomalous conditions such as abnormally low winter snow cover, spring heat waves, and clear summer skies were identified as critical contributors to the unprecedented levels of ice melt. With this insight, we designed our methodology to use the correlation score as a metric to cluster the input features. This technique enhances the representational capacity of the model by prioritizing the intra-cluster relationships. We then apply the methodology to identify anomalous periods across various Arctic regions, we further identified features contributing mostly to these anomalies. Northwest (NW), Northeast (NE), Southwest (SW), and Southeast (SE) are the selected regions of interest in this study due to the availability of ground-truth data.

Our methodology achieved a significantly higher F1-score on unseen data of a benchmark anomaly detection dataset \cite{deng2021graph, su2019robust}. This highlights the potential of our approach to contribute significantly to the field of anomaly detection.

To summarize, the main contributions of this research are:
\vspace{-0.2in}
 \begin{itemize}
    \item We propose a Cluster-VAE framework for anomaly detection in climate data.
    \vspace{-0.1in}
    \item We design a dynamic threshold algorithm characterized by climate attributes such as seasonality to localize the anomaly detection phase.
    \vspace{-0.1in}
    \item We provide a practical tool for identifying anomalous periods and features in climate studies.
\end{itemize}

The next section describes related work on anomaly detection. Section 3 introduces our methodology and framework for anomaly detection, while Section 4 analyzes our experiment and results. Finally, Section 5 summarizes the outcome of our approach and discusses the broader impact of the method.


\vspace{-0.2in}
\section{Related Work}\label{rwork}
\vspace{-0.1in}
\subsection{Anomaly Detection}
\vspace{-0.1in}
Time-series anomaly detection, often a paramount task in data analysis, has garnered significant attention in the literature due to its wide applicability in areas such as network security, weather forecasting, and fault diagnosis. Broadly, the literature on anomaly detection can be classified into two primary strategies:

\begin{enumerate}
\vspace{-0.1in}
\item \textbf{Univariate Anomaly Detection:} This approach targets single time series data. It aims to identify anomalies based solely on individual timestamp deviations from established norms or patterns \cite{ren2019time,malhotra2015long,ishimtsev2017conformal}. While this method is effective for isolated datasets where individual time series aberrations are of primary concern, it often fails to capture the essence of real-world systems. 
\vspace{-0.2in}
\item \textbf{Multivariate Anomaly Detection:} Recognizing the limitations of the univariate approach in scenarios with interlinked time series, the multivariate technique emerged as a more encompassing strategy. It doesn't merely consider the temporal dynamics within an individual time series. Instead, it holistically captures the inter-dependencies across multiple time series. By modeling these interconnected series as a singular entity, the approach can identify anomalies that manifest in coordinated deviations across various time series \cite{hundman2018detecting,su2019robust,zong2018deep}. However, interpreting the anomalies can be challenging. 
\end{enumerate}

Traditional clustering algorithms such as K-Means \cite{yu2018two}, and DBSCAN clustering have long been cornerstone techniques for anomaly detection. These algorithms operate under the foundational hypothesis that normal data points coalesce into dense regions in the feature space, leaving anomalies to lie in sparse, less populated areas. However, in the context of high-dimensional time series data, conventional clustering methodologies face intrinsic challenges. A pivotal constraint pertains to the retention of the inherent structure of the time series data, a concern echoed by \cite{LI2021106919}. 

Graph-based clustering methodologies have also been proffered as alternative paradigms to traditional methods. For example, \cite{li2021dynamic} presented a Dynamic Graph Embedding-based (DynGPE) model aimed at clustering climatic events. While this methodology exhibits potential for capturing intricate relationships among data points, it has limitations when it comes to preserving temporal dependencies inherent in the dataset, an aspect that cannot be trivialized in certain application domains like climatology.

These works summarize the several approaches that have been employed to tackle the challenges and limitations of clustering-based anomaly detection, particularly in multivariate time series data. However, as elucidated, the challenges of preserving intricate temporal and multivariate dependencies remain only partially addressed. 

\textit{In light of the limitations observed in existing methodologies, we propose a novel approach aimed at providing a more holistic solution that enhances the models' performance, retains the temporal dependencies, and identifies the anomalous features.}

\vspace{-0.2in}
\section{Methods}\label{meth}
\vspace{-0.1in}
\subsection{Definitions}
\vspace{-0.08in}
\begin{definition}
    Multivariate time series (MTS): A MTS consists of multiple univariate time series \(t\) of a finite sequence of values with \(m\) unique observations $t$ = \{$1,...,m$\}. MTS $T$ = \{$t_1,...t_{|T|}$\}, $|T|$ is the number of time series.
\end{definition}

\vspace{-0.1in}
\begin{definition}
    Anomaly scores $Y$ is a combination of the reconstruction error and the KL-divergence. 
    \begin{equation}
        Y(t^i) = \lambda_1*x^i + \lambda_2 * KL(P||Q)
    \end{equation}
    where $x^i$ is the reconstruction error at the i-th point, $KL(P||Q)$ is the KL-divergence between two probability distributions $P$ and $Q$, while $\lambda_1$ and $\lambda_2$ are hyperparameters that weigh the importance of the reconstruction error and KL-divergence. 
\end{definition}

\vspace{-0.2in}
    
\subsection{Problem Formulation}
\vspace{-0.1in}
For each region, if $t_i$ and $t_j$ are time series, we aim to enhance representation sampling of $t_i$ and $t_j$ in the latent space, by prioritizing the intra-cluster relationship to minimize the reconstruction loss. We aim to identify temporal anomalies in the region and estimate features significantly influencing anomalies.
\vspace{-0.15in}
\subsection{Framework}
\vspace{-0.1in}
The framework of our model as illustrated in Figure \ref{fig1} comprises of the following steps:
\vspace{-0.1in}
\begin{figure}
    \centering
    \includegraphics[width=0.9\columnwidth]{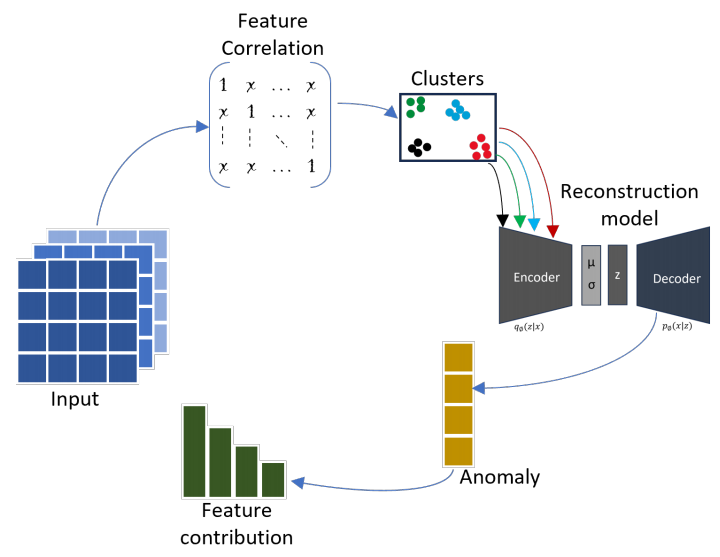}
    \caption{Model Framework}
    \label{fig1}
        \vspace{-0.2in}
\end{figure} 

\begin{itemize}
\setlength{\itemsep}{-0.02in}
    \item We obtain the correlation score for the MTS using Pearson's correlation coefficient. 
    Given $n$ variables, the correlation matrix $R$ would be an $n$ x $n$ matrix. For $n$ variables: $t_1, t_2,..., t_n$, the Pearson correlation coefficient between $t_i$ and $t_j$ (where $i$ and $j$ are between $1$ and $n$) is given by:
    \begin{equation}
        r_{ij} = \frac{\sum(t_i - \Bar{t}_i)(t_j - \Bar{t}_j)}{\sqrt{\sum(t_i - \Bar{t}_i)^2 \sum(t_j - \Bar{t}_j)^2}}
        \label{eq2}
    \end{equation}
 where $\Bar{t}_i$ and $\Bar{t}_j$ are the means of variables $t_i$ and $t_j$. 
Note that \( r_{ij} = r_{ji} \)) because the correlation between \( t_i \) and \( t_j \) is the same as the correlation between \( t_j \) and \( t_i \).

    \item We then cluster the correlated variables using K-means, here the variables are vectors and the K-means algorithm clusters these vectors based on their similarities. For each variable $t_i$ assign to cluster $C_j$ is represented in equation 3
    \begin{equation}
        C_j = argmin_{c_k}d(t_i, c_k)
    \end{equation}
   $c_k$ represents the k-th centroid, and $d$ is the distance measure $=\sqrt{2(1 - r)}$ Euclidean distance. $r$ is the Pearson correlation coefficient. The optimal number of clusters was estimated using the Elbow method and evaluated using Silhouette coefficient score.

   \item Each cluster $C_1, C_2,...C_{|j|}$ having variables $\in$ ($t_i$ and $t_j$) where $C_{|j|}$ is the number of clusters, are fed into a reconstruction based model in parallel. We implemented the LSTM network to learn the temporal dependencies in the input and adopt the VAE architecture \cite{an2015variational} for the reconstruction based model.

   \item We adopted the model loss (Equation \ref{Lvae}) as our anomaly score and then identified the anomalies using a Dynamic Threshold algorithm based on POT. The identified anomalies are then used as labels to identify features significantly influencing extreme events in each region using the feature perturbation method.
   \begin{equation}
   \footnotesize
    Loss_{VAE} = -E_{q\phi(z|x^i)}[logp_\theta(x^i|z)] + KL(q_\phi(z|x^i)||p_\theta(z)) 
    \label{Lvae}
    \end{equation}

\end{itemize}
\vspace{-0.3in}
\subsection{Data Preprocessing}
\vspace{-0.1in}
Reconstruction-based models can be vulnerable to irregular and abnormal instances present in training data. To mitigate this, we adopted the Interquartile Range (IQR) method—a classical approach in univariate anomaly detection. By leveraging IQR, we identified and flagged abnormal timestamps within each individual time series from the training set. For each abnormal instance, its value was replaced by computing the mean of its neighboring normal data points. We reshape our data into $D$ x $T$ x $F$ dimensions, where $D$ = $samples$, $T$ = $timesteps$, and $F$ = $features$, to reshape the data, a statefull rolling window was applied to the training and testing data with a window size of 14 timesteps.
\vspace{-0.15in}
\subsection{Anomaly Detection via Dynamic Thresholding (DT) with POT}
\vspace{-0.1in}

We identified characteristics of the dataset such as seasonality, trend, or periodicity, and then based on these characteristics, we defined a sliding window mechanism. We segment the anomaly score of the time series into three segments, with the time steps in the first and last segment equivalent to the window size, this allows for handling of edge cases. Our procedure commences with the extraction of an initial anomaly threshold for the first segment, leveraging the POT \cite{siffer2017anomaly} methodology. This serves as a benchmark to categorize each temporal instance within this segment as either normal or anomalous. Next, we move the sliding window by a magnitude equivalent to half its span. Within this shifted window, we recompute the threshold via POT. If the data points from the initial time steps surpass the computed threshold, they are marked anomalous. This half-length transition strategy for the sliding window is meticulously chosen to ensure that the internal sequence distribution within the window is not affected by the identified anomalous time steps, while simultaneously enabling a localized threshold estimation based on neighboring temporal values.

\vspace{-0.2in}
\section{Experiments and Results}
\vspace{-0.1in}
\subsection{Datasets and Evaluation Metrics}
\begin{figure}[!t]
    \centering
    \includegraphics[width=0.47\linewidth]{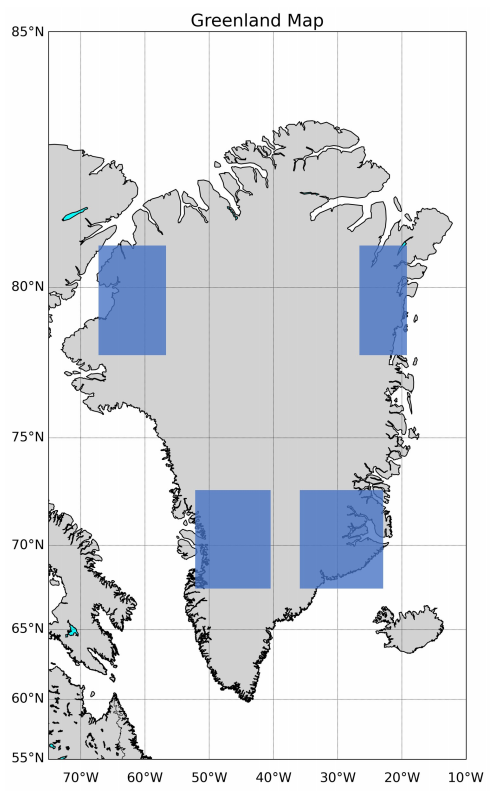}\vspace{-0.1in}
    \caption{Greenland map with area studied shaded in blue}
    \label{greenland}
    \vspace{-0.15in}
\end{figure}

\vspace{-0.1in}
This section delineates the datasets and performance metrics employed in our experimental evaluation. A summary of the datasets' characteristics is presented in Table \ref{features}, and they are further elaborated below:
\vspace{-0.08in}
\begin{itemize}
    \item \textbf{ERA5 Dataset:} The ERA5 dataset is the fifth generation of ECMWF atmospheric reanalyses. In this research, we analysed 13 features with periods between 1941 and 2022. The regions studied are shaded blue in Figure \ref{greenland}.
        \vspace{-0.08in}
\end{itemize}

\begin{table}[!t]
\centering
\caption{Climate features used in the study}
\begin{adjustbox}{max width=0.7\columnwidth}
\begin{tabular}{lllll}
\hline
Sub-domain                   & Features                                       \\\hline
\multirow{2}{*}{Temperature} & 2m temperature $t2m$                         \\
                             & Skin temperature $skt$                         \\ \hline
\multirow{2}{*}{Pressure}    & Mean sea level pressure $msl$                  \\
                             & Surface pressure $sp$                          \\ \hline
\multirow{2}{*}{Wind}        & 10m u-component of wind $u10$                  \\
                             & 10m v-component of wind $v10$                  \\ \hline
\multirow{2}{*}{Radiation}   & Surface solar radiation downwards $ssrd$      \\
                             & Surface thermal radiation downwards   $strd$   \\ \hline
Clouds                       & Total cloud cover $tcc$                        \\ \hline
Precipitation                & Total precipitation $tp$                      \\ \hline
\multirow{3}{*}{Snow}        & Snowmelt $smlt$                                \\
                             & Snow albedo $asn$                              \\
                             & Snow depth $sd$                              \\ \hline 
\end{tabular}
\label{features}
    \vspace{-0.15in}
    \end{adjustbox} \vspace{-0.15in}
\end{table} 

To demonstrate the effectiveness of our method, we evaluated its performance on two benchmark datasets: WADI (Water Distribution Dataset) \cite{ahmed2017wadi} and SMD (Sever Machine Dataset) \cite{su2019robust}. We use precision, recall, and F1-Score as metrics over the test dataset label. The result in Table \ref{base} shows the performance of our method compared to baseline models.

\begin{table}[!t]
    \centering
    \caption{Model accuracy in terms of precision, recall, and f-1 score from multiple models. The results for the DAGMM, LSTM-VAE, MAD-GAN, GDN, OmniAnomaly are as reported in \cite{deng2021graph, su2019robust}}
    \begin{adjustbox}{max width=1\columnwidth}
    \begin{tabular}{lrrr|rrr}
    \hline
        \multirow{2}{*}{Method} & \multicolumn{3}{c|}{WADI} & \multicolumn{3}{c}{SMD} \\ 
        
                                 & Prec & Rec & F1 & Prec & Rec & F1 \\ 
    \hline
        DAGMM                   & 0.54 & 0.27 & 0.36 & 0.59 & \textbf{0.87} & 0.70 \\ 
        LSTM-VAE                & 0.88 & 0.15 & 0.25 & 0.79 & 0.70 & 0.78 \\ 
        MAD-GAN                 & 0.41 & 0.34 & 0.37 & -    & -    & -    \\ 
        GDN                     & \textbf{0.98} & \textbf{0.40} & \textbf{0.57} & -    & -    & -    \\ 
        OmniAnomaly             & -    & -    & -& \textbf{0.83} & \textbf{0.94} & \textbf{0.88}    \\ 
    \hline 
        \textbf{Cluster-LSTM-VAE with DT} & \textbf{0.96} & \textbf{0.78} & \textbf{0.87} & \textbf{0.97} & 0.75 & \textbf{0.85} \\ 
    \hline
    \end{tabular}
    \end{adjustbox}
    \label{base}
        \vspace{-0.15in}
\end{table}

\begin{table}[!t]
    \centering
    \caption{Original feature clusters obtained for each region (Northeast NE, Northwest NW, Southwest SW, and Southeast SE Greenland) without removing outliers}
    \begin{adjustbox}{max width=1\columnwidth}
    \begin{tabular}{llll}
    \hline
        \textbf{NE Clusters} & \textbf{NW Clusters} & \textbf{SW Clusters} & \textbf{SE Clusters} \\ 
        \hline
        u10,v10,asn,sd,smlt & u10,v10,smlt,tcc,tp & u10,v10,smlt,tcc,tp & u10,smlt \\ 
        t2m,skt,ssrd,strd & t2m,skt,ssrd,strd & t2m,skt,strd & v10,strd,tcc,tp \\ 
        msl,sp & msl,sp & msl,sp,ssrd & t2m,msl,skt,sp,ssrd \\ 
        tcc,tp & asn,sd & asn,sd & asn,sd \\ \hline
    \end{tabular}
    \end{adjustbox}
    \label{ERA5_Cluster}
    \vspace{-0.15in}
\end{table}

\begin{table}[!t]
    \centering
    \caption{Feature clusters obtained for each region (Northeast NE, Northwest NW, Southwest SW, and Southeast SE Greenland) after removing outliers}
    \begin{adjustbox}{max width=1\columnwidth}
    \begin{tabular}{llll}
    \hline
        \textbf{NE Clusters} & \textbf{NW Clusters} & \textbf{SW Clusters} & \textbf{SE Clusters} \\ 
        \hline
        u10,asn,sd & u10,v10,smlt,tcc,tp & u10,v10,tcc,tp & u10,asn,sd \\ 
        v10,msl,sp & t2m,skt,ssrd,strd & t2m,skt,strd & v10,strd,tcc,tp \\ 
        t2m,skt,smlt,ssrd,strd & msl,sp & msl,sp,ssrd & t2m,skt,smlt,ssrd \\ 
        tcc,tp & asn,sd & asn,sd,smlt & msl,sp \\ \hline
    \end{tabular}
    \end{adjustbox}
    \label{ERA5_Cluster2}
        \vspace{-0.05in}
\end{table}
\vspace{-0.2in}
\subsection{Results on ERA5 Data}
\vspace{-0.06in}
Table \ref{ERA5_Cluster2} is the expected normal cluster obtained after removing outliers from individual feature using IQR and Table \ref{ERA5_Cluster} is the clusters under the influence of outliers. These clusters indicate the varied dynamics inherent in each region. A comprehensive analysis of these dynamics will be the focus of a forthcoming publication. In the current study, we utilize the clusters in Table \ref{ERA5_Cluster2} as input for our model.
\begin{figure}[t!]
    \centering
    \includegraphics[width=1\linewidth]{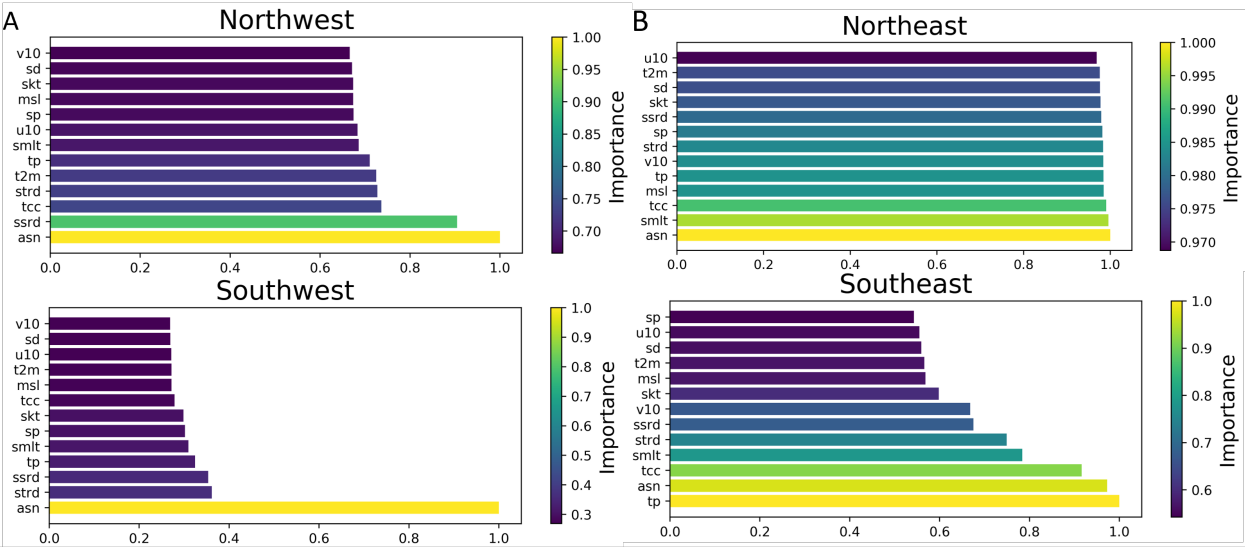}\vspace{-0.1in}
    \caption{The feature importance ranking computed using feature perturbation. A) 2019, B) 2021} 
    \label{importance}
        \vspace{-0.1in}
\end{figure}

\begin{figure}[t!]
    \centering
    \includegraphics[width=1\linewidth]{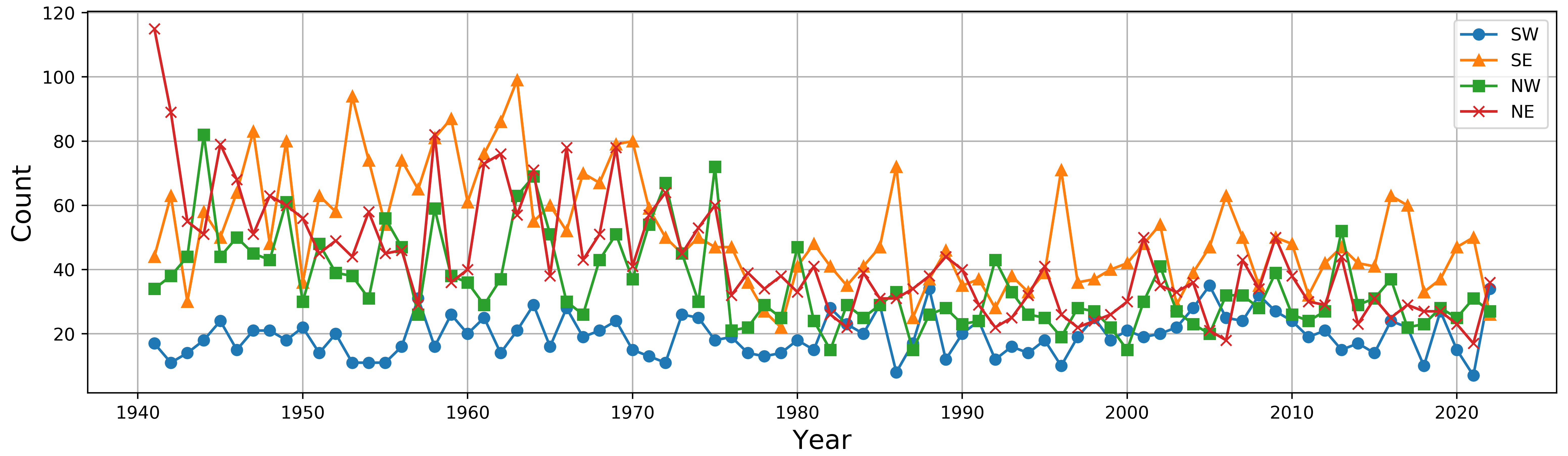}\vspace{-0.1in}
    \caption{The year to year trend of observed anomalies from 1941 to 2022}
    \label{trend}
    \vspace{-0.2in}
\end{figure}

\vspace{-0.2in}
\subsection{Ground Truth Validation}
\vspace{-0.1in}
Figure \ref{importance} shows the ranking of the anomalous features that influenced extreme snowmelt events during specific period. In 2019, the Snow Albedo ('asn') emerged as the most impactful feature in Western Greenland's unprecedented snowmelt, as shown in Figure \ref{importance}A. Concurrently, Surface Solar Radiation Downward ('ssrd') was notably influential in the Northwest. These findings align with the National Snow and Ice Data Center (NSIDC) 2019 report \cite{nsidc}, which attributed the extensive melt to a combination of low winter snow cover, a heatwave, and a sunny summer, leading to a decrease in albedo and, consequently, enhanced melting. In 2021, despite a typical melt season, an unprecedented rainfall at the Summit Station led to significant snowmelt especially in Southern Greenland. This event reduced snow albedo, thereby increasing surface melting \cite{nsidc2021, thoman2022arctic}, corroborating the significance of Total Precipitation ('tp') and Snow Albedo ('asn') in the Southeast, as indicated in Figure \ref{importance}B.

The findings from our proposed framework harmonize seamlessly with the ground truth data \cite{nsidc, nsidc2021, thoman2022arctic}, it identifies 'asn' as the predominant feature across all examined regions and periods. Notably, the absence of significant features in the Northeast in 2021 implies a lack of exceptional climatic events in that region for the year. Further analysis of the anomalous trend from 1941 to 2022 (Figure \ref{trend}) reveals a dip in anomalies between 1975 and 1980, and the Southeast consistently exhibited more anomalies, topics we plan to explore in future publications. These observations, coupled with external climate data, provide a comprehensive understanding of the temporal and spatial dynamics of Arctic snowmelt.

    \vspace{-0.3in}
\section{Conclusion}
    \vspace{-0.1in}  
In this study, we offer a novel perspective on the critical drivers of extreme snowmelt events in Greenland. We proposed our model Cluster-LSTM-VAE with Dynamic Thresholding technique for identifying the critical drivers, such as snow albedo, solar radiation, and total precipitation, that significantly influence these events. Our findings not only provide valuable insights into the region-specific climatic dynamics but also set a foundation for future research focused on understanding and predicting snowmelt patterns in response to global climate change. The identification of these key features is instrumental in informing policy decisions and mitigation strategies aimed at preserving Arctic environments.
\section{Acknowledgments}
This work is funded by the National Science Foundation (NSF) Award \#2118285, "HDR Institute: HARP-Harnessing Data and Model Revolution in the Polar Regions". The WADI dataset was provided by iTrust, Center for Research in Cyber Security, Singapore University of Technology and Design.

\bibliographystyle{IEEEbib}

\begin{thebibliography}{10}

\bibitem{deng2021graph}
Ailin Deng and Bryan Hooi,
\newblock ``Graph neural network-based anomaly detection in multivariate time series,''
\newblock in {\em Proceedings of the AAAI conference on artificial intelligence}, 2021, vol.~35, pp. 4027--4035.

\bibitem{su2019robust}
Ya~Su, Youjian Zhao, Chenhao Niu, Rong Liu, Wei Sun, and Dan Pei,
\newblock ``Robust anomaly detection for multivariate time series through stochastic recurrent neural network,''
\newblock in {\em Proceedings of the 25th ACM SIGKDD international conference on knowledge discovery \& data mining}, 2019, pp. 2828--2837.

\bibitem{ren2019time}
Hansheng Ren, Bixiong Xu, Yujing Wang, Chao Yi, Congrui Huang, Xiaoyu Kou, Tony Xing, Mao Yang, Jie Tong, and Qi~Zhang,
\newblock ``Time-series anomaly detection service at microsoft,''
\newblock in {\em Proceedings of the 25th ACM SIGKDD international conference on knowledge discovery \& data mining}, 2019, pp. 3009--3017.

\bibitem{malhotra2015long}
Pankaj Malhotra, Lovekesh Vig, Gautam Shroff, Puneet Agarwal, et~al.,
\newblock ``Long short term memory networks for anomaly detection in time series.,''
\newblock in {\em Esann}, 2015, vol. 2015, p.~89.

\bibitem{ishimtsev2017conformal}
Vladislav Ishimtsev, Alexander Bernstein, Evgeny Burnaev, and Ivan Nazarov,
\newblock ``Conformal $ k $-nn anomaly detector for univariate data streams,''
\newblock in {\em Conformal and Probabilistic Prediction and Applications}. PMLR, 2017, pp. 213--227.

\bibitem{hundman2018detecting}
Kyle Hundman, Valentino Constantinou, Christopher Laporte, Ian Colwell, and Tom Soderstrom,
\newblock ``Detecting spacecraft anomalies using lstms and nonparametric dynamic thresholding,''
\newblock in {\em Proceedings of the 24th ACM SIGKDD international conference on knowledge discovery \& data mining}, 2018, pp. 387--395.

\bibitem{zong2018deep}
Bo~Zong, Qi~Song, Martin~Renqiang Min, Wei Cheng, Cristian Lumezanu, Daeki Cho, and Haifeng Chen,
\newblock ``Deep autoencoding gaussian mixture model for unsupervised anomaly detection,''
\newblock in {\em International conference on learning representations}, 2018.

\bibitem{yu2018two}
Shyr-Shen Yu, Shao-Wei Chu, Chuin-Mu Wang, Yung-Kuan Chan, and Ting-Cheng Chang,
\newblock ``Two improved k-means algorithms,''
\newblock {\em Applied Soft Computing}, vol. 68, pp. 747--755, 2018.

\bibitem{LI2021106919}
Jinbo Li, Hesam Izakian, Witold Pedrycz, and Iqbal Jamal,
\newblock ``Clustering-based anomaly detection in multivariate time series data,''
\newblock {\em Applied Soft Computing}, vol. 100, pp. 106919, 2021.

\bibitem{li2021dynamic}
Gen Li and Jason~J Jung,
\newblock ``Dynamic graph embedding for outlier detection on multiple meteorological time series,''
\newblock {\em Plos one}, vol. 16, no. 2, pp. e0247119, 2021.

\bibitem{an2015variational}
Jinwon An and Sungzoon Cho,
\newblock ``Variational autoencoder based anomaly detection using reconstruction probability,''
\newblock {\em Special lecture on IE}, vol. 2, no. 1, pp. 1--18, 2015.

\bibitem{siffer2017anomaly}
Alban Siffer, Pierre-Alain Fouque, Alexandre Termier, and Christine Largouet,
\newblock ``Anomaly detection in streams with extreme value theory,''
\newblock in {\em Proceedings of the 23rd ACM SIGKDD international conference on knowledge discovery and data mining}, 2017, pp. 1067--1075.

\bibitem{ahmed2017wadi}
Chuadhry~Mujeeb Ahmed, Venkata~Reddy Palleti, and Aditya~P Mathur,
\newblock ``Wadi: a water distribution testbed for research in the design of secure cyber physical systems,''
\newblock in {\em Proceedings of the 3rd international workshop on cyber-physical systems for smart water networks}, 2017, pp. 25--28.

\bibitem{nsidc}
``Large ice loss on the greenland ice sheet in 2019,''
\newblock {\em Greenland Ice Sheet Today}, 2019.

\bibitem{nsidc2021}
``Greenland surface melting in 2021,''
\newblock {\em Greenland Ice Sheet Today}, 2021.

\bibitem{thoman2022arctic}
Richard~L Thoman, Matthew~L Druckenmiller, Twila~A Moon, LM~Andreassen, E~Baker, Thomas~J Ballinger, Logan~T Berner, Germar~H Bernhard, Uma~S Bhatt, Jarle~W Bjerke, et~al.,
\newblock ``The arctic,''
\newblock {\em Bulletin of the American Meteorological Society}, vol. 103, no. 8, pp. S257--S306, 2022.

\end{thebibliography}

\begingroup
\renewcommand{\thefootnote}{}
\footnotetext{2024 IEEE International Geoscience and Remote Sensing Symposium, IGARSS 2024, 07-12 July 2024, Athens, Greece.}
\endgroup

\end{document}